\RequirePackage[T1]{fontenc}
\RequirePackage{graphicx}
\RequirePackage{algorithm}
\RequirePackage{algpseudocode}
\RequirePackage{amsmath}
\RequirePackage{mathtools}
\RequirePackage{array}
\RequirePackage{makecell}
\RequirePackage{boldline} 
\RequirePackage{booktabs}

\newcolumntype{L}[1]{>{\PreserveBackslash\raggedright}p{#1}}
\newcolumntype{C}[1]{>{\PreserveBackslash\centering}p{#1}}

\documentclass[runningheads]{llncs}

\usepackage{microtype}
\begin{document}

\title{RIFF: Inducing Rules for Fraud Detection from Decision Trees}

\titlerunning{RIFF}
%

\author{Lucas Martins\inst{1} \and
João Bravo\inst{1} \and
Ana Sofia Gomes\inst{1} \and
Carlos Soares\inst{2} \and
Pedro Bizarro\inst{1}}
\authorrunning{L. Martins et al.}
%
\institute{Feedzai, Portugal \and
Faculdade de Engenharia da Universidade do Porto, Portugal
}
\maketitle              
\begin{abstract}

Financial fraud is the cause of multi-billion dollar losses annually.
Traditionally, fraud detection systems rely on rules due to their transparency and interpretability, key features in domains where decisions need to be explained.
However, rule systems require significant input from domain experts to create and tune, an issue that rule induction algorithms attempt to mitigate by inferring rules directly from data. 
We explore the application of these algorithms to fraud detection, where rule systems are constrained to have a low false positive rate (FPR) or alert rate, by proposing RIFF, a rule induction algorithm that distills a low FPR rule set directly from decision trees. Our experiments show that the induced rules are often able to maintain or improve performance of the original models for low FPR tasks, while substantially reducing their complexity and outperforming rules hand-tuned by experts.

\end{abstract}

\section{Introduction}

Despite the advent of modern machine learning (ML) algorithms, rule systems continue to be important in many domains~\cite{Aparicio2020ARMS,yan2022FIGS}. Their simplicity and interpretability, often requirements in high stake problems, as well as their longstanding presence has earned the trust of many financial institutions. Many continue to use rule systems as their only solution for fraud detection, while others use them alongside machine learning models.

However, building rule sets traditionally requires expert input and their predictive performance is typically worse than modern machine learning models. This could potentially be attributed, at least in part, to the fact that rules are not automatically inferred from data, but instead manually created and tuned.

While there are several induction algorithms that infer rules from data \cite{quinlan1986ID3,Breiman1984CART,quinlan1993C45,yan2022FIGS,litao2019DRNet,kusters2022R2N,michalski1986AQ,cendrowska1987prism,clark1989CN2}, applying them to fraud detection can be problematic due to the extreme class imbalance that is often present, and the requirement to have very low FPR values, typically under 2\%. The latter is necessary as incorrectly flagging legitimate transactions can cause friction, eroding customer trust, leading to financial losses, and putting undue pressure on manual reviewers. For this reason, experts try to minimize false positives when considering the trade-off with false negatives. Another requirement is to induce rules that are easily understood by experts, for two reasons: firstly, experts often need to review the decision made by a rule, and, as such, they must understand the reason behind it; secondly, experts need to manually modify rules periodically to keep up with new fraud patterns. As such, a rule system ideally has as few rules as possible so that it can be better understood by humans.

Our main contribution is a rule induction algorithm, RIFF, that leverages decision trees to build low FPR rule sets for fraud detection. We benchmark RIFF against state-of-the-art decision trees algorithms, CART and FIGS, and against expert made rules. Our experiments use both publicly available and private real world transaction data, and we benchmark RIFF using trees generated by CART, FIGS, and by our own modified version of FIGS, FIGU, that aims to reduce overlap between rules induced from different trees.

\section{Related Work}~\label{sec:related_work}

Prior work on rule set induction can be divided into two distinct approaches. \emph{Separate-and-conquer algorithms}, also known as covering algorithms, form rule sets by adding rules one by one until a stopping criterion is met~\cite{Michalski1980PatternRA,Michalski1986TheMI,cendrowska1987prism,quinlan1990FOIL,cohen1995ripper,furnkranz1994IREP}. They build or refine rules incrementally, typically relying on a heuristic to choose the best condition or rule to add. In each iteration, these algorithms remove examples covered by the rule set so that new rules can focus on data that is not covered by the current rule set.

On the other hand, \emph{divide-and-conquer algorithms} such as ID3~\cite{quinlan1986ID3}, C4.5~\cite{quinlan1993C45} and CART~\cite{Breiman1984CART}, use decision trees to describe the data. These trees are grown in a greedy fashion, by iteratively splitting a current leaf node based on the value of one attribute to maximize a chosen criterion, such as \emph{information gain}.
FIGS~\cite{yan2022FIGS} expands on these algorithms (namely CART) by introducing the option of adding a new tree by splitting on a new root node instead of an existing leaf node. Each tree independently contributes to the model with a score that is summed to produce the final prediction.

Our work leverages both of these approaches. RIFF employs \emph{divide-and-conquer} methods to build decision trees, and it uses a \emph{separate-and-conquer} heuristic to build a rule set with the best performing rules while enforcing a low FPR or alert rate constraint.

\section{Rule Induction for Fraud Detection}
\label{sec:riff}

Rule systems used in the context of fraud detection are typically composed of tens to hundreds of simple rules, each designed to capture a particular fraud pattern. These rules are usually a conjunction of a small set of logical conditions that can be understood by a human and evolve with time by tuning specific thresholds. Fraud detection systems are also usually constrained to operate with an overall low False Positive Rate (FPR) or Alert Rate (AR). This is to limit the friction caused to legitimate users or to limit the total number of alerts generated according to the capacity of a fraud analyst team.

\begin{figure}[h]
    \centering
    \includegraphics[width=0.95\linewidth]{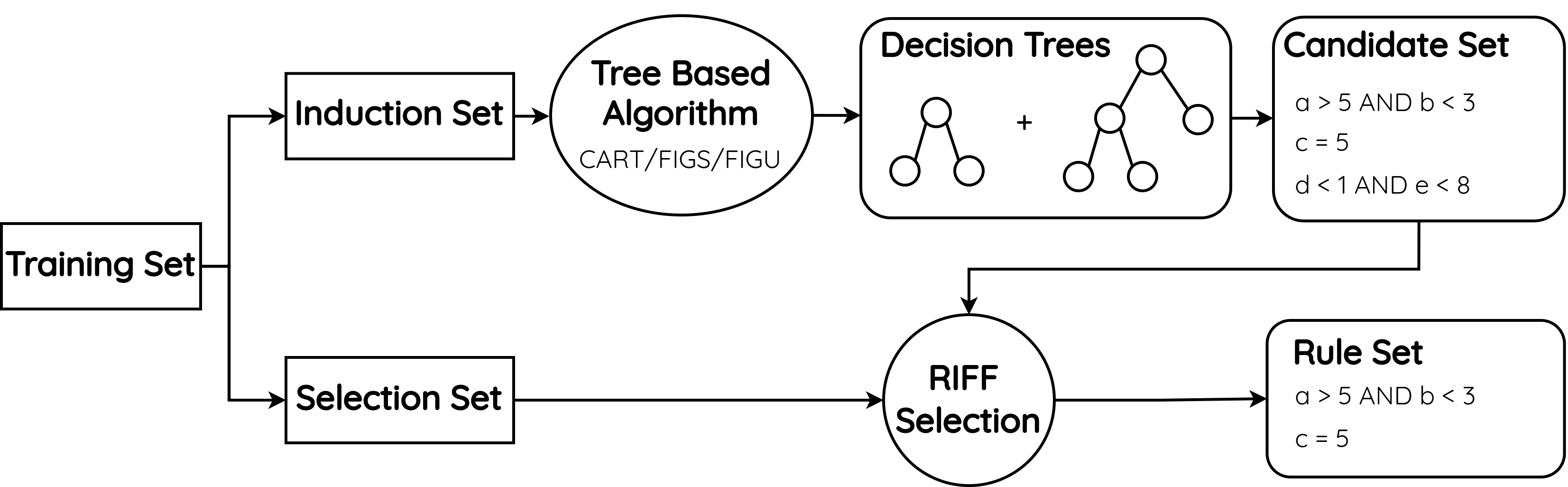}
    \caption{RIFF Overview}
    \label{fig:riff_overview}
\end{figure}

Decision tree algorithms like CART have formed the basis of state-of-the-art algorithms for tabular data when used in ensembles such as Random Forests \cite{breiman2001RandomForests} or Gradient-Boosted Decision Trees \cite{friedman2001greedy}. However, even single decision trees can be hard to understand and interpret by humans and they can't be manually tuned by experts.
We thus propose leveraging these algorithms to generate candidate rules with good discriminative performance.
For this, our proposed algorithm, RIFF, is split into two steps (see Figure~\ref{fig:riff_overview}):
\begin{enumerate}
    \item We induce a set of candidate rules from the leaves of a tree-based model trained on an \emph{induction set}. This candidate set would, ideally, contain different rules with high precision, corresponding to leaves with high purity of fraud examples. 
    \item We greedily select the rules with highest precision from the candidate set, based on their performance on an out of sample \emph{selection set}.

\end{enumerate}
Over the next sections we describe in detail these two steps.

\subsection{Rule Induction from Decision Trees}\label{sec:rule_induction_cart}

In order to generate a low FPR rule set, we extract rules from decision trees. We base this decision on the fact that the typical splitting criterion tries to find the \emph{purest} leaves, i.e., leaves with highest precision that in theory maximize the amount of gained recall per FPR. For this reason, we will assume that all rules in the extracted candidate set predict the \emph{positive} class.

\begin{figure}[ht]
    \centering
    \includegraphics[width=0.8\textwidth]{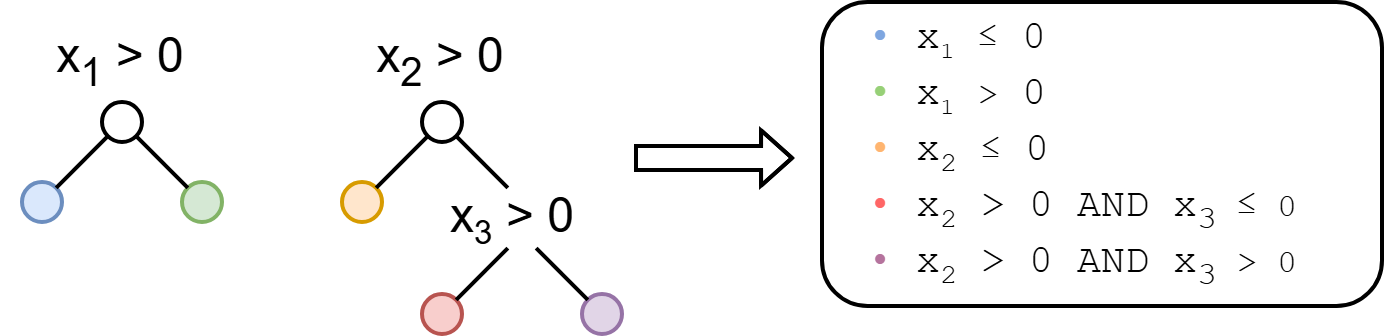}
    \caption{Extracting rules from a FIGS model}
    \label{fig:figs_trees}
\end{figure}

After creating a tree with suitable, high purity leaves, we form a candidate rule set by extracting one rule for each leaf. This is accomplished by traversing the path from the root node to each leaf, forming a new rule as a conjunction of the conditions in this path. Figure~\ref{fig:figs_trees} shows an example where a tree model with 5 leaves was converted to a rule set.

This method does not extend well to additive tree models like FIGS, because it ignores tree scores when converting leaves to decision rules. To choose the best split on a tree $i \in \{1, 2, \ldots, T\}$, where $T$ is the number of trees in the model, FIGS uses a mean squared error criterion with residuals calculated by subtracting from the label the predictions from all other trees $j$ as targets. For a sample $(\mathbf{x}, y)$, the residual, $r_i$, for tree $i$ is thus given by:
\begin{equation}
	r_i(\mathbf{x}, y) = y - \sum\limits_{\substack{j=1 \\ j\neq i}}^T \hat{y}_{j}(\mathbf{x})\;,
\end{equation}
\noindent
where $\hat{y}_j(\mathbf{x})$ is the prediction for tree $j$, given by the positive rate for the leaf node where the sample, $\mathbf{x}$, falls. This additive approach means that leaves generated by FIGS may not be pure enough to yield low FPR rules since they are meant to complement the predictions made by other trees.

For this reason, we modify FIGS by binarizing the residual computation, thus turning its Greedy Tree Sums into Greedy Tree Unions (FIGU). Concretely, when considering how to split a current leaf node in a tree $i$, we discard any samples that fall into the support of that node if they are already covered by a current leaf node of another tree $j$. I.e., we discard a sample $(\mathbf{x}, y)$ when evaluating the splitting criterion if: 
\begin{equation}
    \bigvee\limits_{\substack{j=1 \\ j\neq i}}^T \hat{y}_{j}(\mathbf{x}) = \texttt{true}\;,
\end{equation}
where $\hat{y}_{j}(\mathbf{x})$ is now a binary prediction for tree $j$ that evaluates to $\texttt{true}$ if: 1) $\mathbf{x}$ falls into the support of a current leaf in tree $j$ with high enough precision (as specified by a user provided threshold); 2) it falls into the support of the current best leaf of tree $j$ as measured by precision.

\subsection{Rule Selection}

As mentioned in Section \ref{sec:riff}, the rule selection step aims to distill a potentially large set of candidate fraud rules into a smaller set, maximizing the number of fraud cases captured, i.e., the True Positive Rate (TPR) of the system, while keeping its FPR or Alert Rate below a given threshold. For concreteness, we will focus on the former constraint, FPR, in the exposition.

Writing as $\text{cov}(R;\mathcal{D})$ the example set covered by a rule set $R$ on dataset $\mathcal{D}$, we have:
\begin{align*}
    \text{TPR}(R) &= \frac{\left\vert\text{cov}(R; \mathcal{D}^+)\right\vert}{|\mathcal{D}^+|}\;, &
    \text{FPR}(R) &= \frac{\left\vert\text{cov}(R; \mathcal{D}^-)\right\vert}{|\mathcal{D}^-|}\;,
\end{align*}
where we denote by $\mathcal{D}^+$ and $\mathcal{D}^-$ the subsets of positive and negative examples respectively. We can thus formalize the rule selection goal as choosing a subset of rules $S$ from a given set of candidate rules $C = \{c_1, c_2, \ldots, c_n\}$ to solve:
\begin{equation}
    \max_{S\in2^C}  \quad \textrm{TPR}(S) \quad \textrm{s.t.} \quad \textrm{FPR}(S) \leq \text{FPR}_\text{max}\;,
\end{equation}
where $\text{FPR}_\text{max}$ is the user-defined maximum desired FPR.
\noindent

Both TPR and FPR are monotone non-decreasing submodular functions and this optimization problem is NP-hard  \cite{DBLP:journals/corr/IyerB13a}. We therefore propose a simple greedy method, described in Algorithm~\ref{alg:algorithm_label}. It iteratively selects rules from $C$ until a stopping criterion is met, which, in our case is when the FPR of the set of selected rules set surpasses $\text{FPR}_\text{max}$ . We assume that this always occurs over the runtime of the algorithm, i.e., that $\textrm{FPR}(C) \geq \text{FPR}_\text{max}$. At each iteration, it selects the candidate rule with the highest precision in the set of samples that are not yet covered by any existing rules, where the precision of a rule, $c$, in a set of samples  $\mathcal{D}$ is defined as:
\begin{equation}
\text{Precision}(c;\mathcal{D}) = \frac{|\text{cov}(c; \mathcal{D^+})|}{|\text{cov}(c; \mathcal{D^+})|+|\text{cov}(c; \mathcal{D^-})|}\;.
\end{equation}

\begin{algorithm}
\caption{Greedy Rule Selection Algorithm}
\label{alg:algorithm_label}
\begin{algorithmic}
\State \textbf{Input:} $( C = { c_1, c_2, \ldots, c_n } );\ \text{FPR}_\text{max};\ \mathcal{D}$

\State \( \mathcal{D}' \leftarrow \mathcal{D} \) \algorithmiccomment{$\mathcal{D'}$ corresponds to all samples not covered by rules in $S$ }
\State \( S \leftarrow \{\} \) \algorithmiccomment{$S$ is the selection set, containing all selected rules }
\State \(i \leftarrow 0\)

\While{\( \text{FPR}(S; \mathcal{D}) < \text{FPR}_\text{max} \)}
    \State \(i \leftarrow i+1\)
    \State \( r_i \leftarrow \arg\max_{c \in C\setminus S}  \text{Precision}(c; \mathcal{D'}) \)
    \State \( \mathcal{D'} \leftarrow \mathcal{D'} \setminus \text{cov}(r_i; \mathcal{D'}) \)
    \State \(S \leftarrow S \cup \left\{r_i \right\}\)
    
\EndWhile

\State \Return \( r_1, \ldots, r_i \)
\end{algorithmic}
\end{algorithm}

This algorithm returns a list of rules in the order they were selected. Defining $S_i \coloneqq \{ r_1, \ldots, r_i\}$ we have that $S_{l-1}$ is guaranteed to satisfy the $\text{FPR}_\text{max}$ constraint, whereas $S_{l}$ may violate it, with $l$ denoting the length of the returned list.

To compare rule sets with different FPR values generated from different candidate sets, we relax our solution set to include randomized rule sets. Instead of a fixed subset $S$ of the candidate set, we output a probability, $\rho(c)\in[0, 1]$ for every rule in $c\in C$ to be selected. For all but the last rule selected by Algorithm \ref{alg:algorithm_label}, i.e., $c\in S_{l-1}$, a probability of 1 is chosen. For the last rule selected, $r_l$, this probability is chosen in order to match the expected FPR with the desired FPR constraint.
Formally, the probability $\rho(c)$ for a rule $c \in C$ is chosen as:
$$
\rho(c) = \left\{
\begin{array}{ll}
    1 & c \in S_{l-1}\\
    \frac{\text{FPR}_\text{max} - \text{FPR}(S_{l-1})}{\text{FPR}(S_{l}) - \text{FPR}(S_{l-1})} &  c = r_l \\
    0 & c \notin S_l
\end{array}
\right. ,
$$
With this in mind, we can interpret the TPR of this randomized rule system as a random variable with an expected value given by:
$$
\text{TPR}(\rho) = (1 - \rho(r_l))\text{TPR}(S_{l-1}) + \rho(r_l) \text{TPR}(S_{l})\,.
$$

\section{Experiments}

We evaluate RIFF on two public classification datasets: BAF~\cite{jesusBAF2022}, a synthetic bank account fraud dataset, and Taiwan credit~\cite{BacheLichman2013UCI}, a credit card default dataset. We also use a private dataset, containing real transaction fraud data, which
we cannot disclose due to privacy and contractual
reasons. A baseline unique to this dataset is a set of rules manually tuned by data scientists allowing us to compare the rules generated by RIFF against rules handcrafted by experts. An overview of the used datasets can be seen in Table \ref{tab:datasets}.

\setlength\tabcolsep{0.1cm}
\begin{table}[h]
\centering
\caption{Dataset Analysis Summary. The train/validation/test splits are time-based for the BAF and Industry datasets and random for Taiwan Credit.}
\begin{tabular}{lccc}
                    & \textbf{BAF}   & \textbf{Industry} & \textbf{Taiwan Credit}   \\ 
    \toprule
    Task            & Account Fraud    & Transaction Fraud & Credit Card Default  \\ 
    Positive rate   & 1\%              & 7\%               & 22\%                \\ 
    \#samples       & 1M        & 3.5M         & 30K        \\ 
    \#features      & 32               & 113               & 25              \\
    Train split & 75\% & 60\%  & 60\% \\
    Validation and test split & 12.5\% & 20\% & 20\%\\
    \bottomrule
\end{tabular}
\label{tab:datasets}
\end{table}
We sample the training set using a parameterized sample ratio into two smaller subsets, the induction and selection set. We use a sample ratio of 10\% for the BAF and Industry datasets, and a sample ratio of 50\% for the Taiwan credit dataset. In this step, we also parameterize the positive rate for the generated subsets, using a positive rate of 30\% for all datasets. After using the induction set to train CART, FIGS and FIGU models, we extract candidate rules from the generated tree models, as described in Section~\ref{sec:rule_induction_cart}. Then, the selection step of the algorithm extracts the best rules from each candidate set according to their performance on the selection set. In this step of the experiment we use a $\text{FPR}_\text{max}$ of 1\%. We use the validation set to tune the total number of splits used when training the decision-tree model, using a line search over the values [10, 20, 30, 40, 50] and we use the test set for the final evaluation of the generated rule set.

We use LightGBM as a strong baseline for predictive performance for two reasons: firstly, because gradient boosted decision trees are a state-of-the-art algorithm for tabular data; secondly, because of their popularity in fraud detection~\cite{cruz2023fairgbmgradientboostingfairness}.  We also report the performance of the best CART and FIGS models trained in the induction step as divide-and-conquer baselines.

\setlength\tabcolsep{0.3cm}
\begin{table}[h]
\centering
\caption{Recall at 1\% FPR in the test split for BAF, Credit and Industry Datasets}
\begin{tabular}{lccc}
& \textbf{BAF} & \textbf{Industry} & \textbf{Taiwan Credit} \\ 
\toprule
LightGBM &
  $0.252$ &
  $0.531$ &
  $0.084$ \\
\cmidrule(r){1-4} 
Expert Rules &
  - &
  $0.158$ &
  -\\
\cmidrule(r){1-4}
 CART &
  $0.160\, \scriptstyle\pm 0.005$ &
  $0.315\, \scriptstyle\pm 0.075$ &
  $0.063\, \scriptstyle\pm 0.009$
   \\
CART + RIFF &
  $\textbf{0.184}\, \scriptstyle\pm 0.006$ &
  $\textbf{0.362}\, \scriptstyle\pm 0.027$ &
  $\textbf{0.139}\, \scriptstyle\pm 0.018$\\
\cmidrule(r){1-4} 
FIGS &
  $\textbf{0.210} \scriptstyle\pm 0.006$ &
  $\textbf{0.394}\, \scriptstyle\pm 0.032$ &
  $0.067\, \scriptstyle\pm 0.016$\\
FIGS + RIFF &
  $0.158\, \scriptstyle\pm 0.016$ &
  $0.311\, \scriptstyle\pm 0.018$ &
  $\textbf{0.136}\, \scriptstyle\pm 0.019$ \\
FIGU + RIFF &
  $0.155\, \scriptstyle\pm 0.010$ &
  $0.382\, \scriptstyle\pm 0.039$ &
  $0.104\, \scriptstyle\pm 0.007$ \\
\bottomrule
\end{tabular}
\label{tab:results_all}
\end{table}

\setlength\tabcolsep{0.3cm}
\begin{table}[ht]
\centering
\caption{Generated Rule set length for BAF, Credit and Industry Datasets}
\begin{tabular}{lccc}
& \textbf{BAF} & \textbf{Industry} & \textbf{Taiwan Credit} \\ 
\toprule
Expert Rules & - & $13.0$ & -\\
CART + RIFF &
  $10.4\, \scriptstyle\pm 3.647$ &
  $17.8\, \scriptstyle\pm 4.087$ &
  $5.2\, \scriptstyle\pm 1.483$ \\
FIGS + RIFF &
  $8.0\, \scriptstyle\pm 1.732$ &
  $9.2\, \scriptstyle\pm 1.304$ &
  $7.6\, \scriptstyle\pm 1.949$ \\
FIGU + RIFF &
  $\textbf{3.6}\, \scriptstyle\pm 0.548$ &
  $\textbf{3.4}\, \scriptstyle\pm 0.548$ &
  $\textbf{1.0}\, \scriptstyle\pm 0.000$ \\
\bottomrule
\end{tabular}
\label{tab:results_len}
\end{table}

We repeat our setup using 5 different seeds for the model training and the sampling of the induction and selection sets. In Table~\ref{tab:results_all} and~\ref{tab:results_len} we report the average performance and average length of generated rule sets of all seeds respectively, as well as the associated standard deviations. Interestingly, using RIFF on CART always improved its performance, a possible indication that CART was overfitting and RIFF reduced this by selecting its best rules. For the dataset with fewer samples, Taiwan Credit, RIFF increased the performance of CART and FIGS significantly, surpassing even LightGBM's performance. FIGU appears to generate rule sets that have similar performance to FIGS with much fewer rules, an indication that FIGU is able to reduce the overlap between generated trees, leading to shorter and, in theory, simpler to understand rule sets.

\section{Conclusion And Future Work}
In this work we propose RIFF, a rule induction algorithm that builds low FPR rule sets for fraud detection by greedily extracting rules from a tree based model like CART or FIGS. We also propose a slight modification to FIGS, FIGU, that aims to lower decision tree complexity so that it can be used by the RIFF selection algorithm to generate shorter rulesets. We perform a study with real world transaction data that shows that RIFF is able to perform better than expert rules. Our experiments show that RIFF is able to maintain the predictive performance of the original models while reducing their complexity by turning them into rules. In addition, when paired with RIFF, FIGU is able to generate rule sets with fewer rules, with similar performance to rules generated by FIGS.

While RIFF effectively generates a more concise and shorter rule set, it might provide complex, lengthier rules. We could expand the candidate set to also consider all nodes, instead of only leaves. This methodology draws a parallel to pruning methods, as this ideally leads RIFF into choosing more general, lower depth nodes in favour of their more specific, children nodes, similar to pruning.

A possible way to generate a more varied and robust rule set could involve extracting rules from all the trained CART and FIGS models into an unique candidate set. Since our setup subsamples the training set into subsests, and uses them to train these models, this is equivalent to applying the RIFF selection algorithm to a Random Forest~\cite{breiman2001RandomForests} or Bagging FIGS~\cite{yan2022FIGS}.


%
%
%

\bibliographystyle{splncs04}
\bibliography{paper.bib}

\end{document}